\documentclass[10pt,twocolumn,letterpaper]{article}

\usepackage{cvpr}
\usepackage{times}
\usepackage{epsfig}
\usepackage{graphicx}
\usepackage{amsmath}
\usepackage{amssymb}

\usepackage{bigstrut}
\usepackage{booktabs}

\usepackage[breaklinks=true,bookmarks=false]{hyperref}
\cvprfinalcopy % *** Uncomment this line for the final submission

\def\cvprPaperID{****} % *** Enter the CVPR Paper ID here

\ifcvprfinal\fi
\begin{document}

%%%%%%%%% TITLE
\title{Adversarial Soft-detection-based Aggregation Network for Image Retrieval}

\author{Jian Xu,
Chunheng Wang,
Cunzhao Shi,
and
Baihua Xiao\\
Institute of Automation, Chinese Academy of Sciences (CASIA)\\
%{\tt\small xujian2015@ia.ac.cn}
%\and
}

\maketitle
%\thispagestyle{empty}

%%%%%%%%% ABSTRACT
\begin{abstract}
   In recent year, the compact representations based on activations of Convolutional Neural Network (CNN) achieve remarkable performance in image retrieval. However, retrieval of some interested object that only takes up a small part of the whole image is still a challenging problem. Therefore, it is significant to extract the discriminative representations that contain regional information of the pivotal small object. In this paper, we propose a novel adversarial soft-detection-based aggregation (ASDA) method free from bounding box annotations for image retrieval, based on adversarial detector and soft region proposal layer. Our trainable adversarial detector generates semantic maps based on adversarial erasing strategy to preserve more discriminative and detailed information. Computed based on semantic maps corresponding to various discriminative patterns and semantic contents, our soft region proposal is arbitrary shape rather than only rectangle and it reflects the significance of objects. The aggregation based on trainable soft region proposal highlights discriminative semantic contents and suppresses the noise of background.

   We conduct comprehensive experiments on standard image retrieval datasets. Our weakly supervised ASDA method achieves state-of-the-art performance on most datasets. The results demonstrate that the proposed ASDA method is effective for image retrieval.

   %In this paper, we propose a novel adversarial soft-detection-based aggregation (ASDA) method free from bounding box annotations for image retrieval. %In order to highlight the certain discriminative pattern of objects and suppress the noise of background, we employ trainable soft region proposals that indicate the probability of interested object and reflect the significance of candidate regions.
   %The soft region proposals are calculated based on semantic maps generated by the adversarial erasing strategy to preserve more discriminative and low-redundancy information.
\end{abstract}

%%%%%%%%% BODY TEXT
\section{Introduction}
%图像检索
In recent years, image retrieval has been researched widely.  Image representation is the pivotal modules in image retrieval, and it has a considerable impact on retrieval performance. The image representation methods based on Scale-Invariant Feature Transform (SIFT)~\cite{sift} and Convolutional Neural Network (CNN)~\cite{feature_map} have received sustained attention over the past decades.

After Lowe et al.~\cite{sift} propose SIFT descriptor, SIFT-based image representation~\cite{binary_embedding,bow,vlad,fv_cvpr,tri_embed,faemb,rvd}
achieves remarkable performance in image retrieval.
As a hand-crafted descriptor based on the histogram of oriented gradients,  SIFT has the significant semantic gap with high-level semantic content of image.
With the development of deep learning, CNN-based image retrieval methods~\cite{off_the_shelf,msop,nc,mr,spoc,rmac,crow,interactive,PWA,SBA,netvlad,fine_tune_1,fine_tune_2,CRN,fine_tune_3,fine_tune_4,delf} become the research focus gradually.

Fully connected layer features of CNN are utilized to generate the compact global representation for image retrieval in some recent works~\cite{off_the_shelf,msop,nc}.
After that, feature maps of convolutional layers are employed to obtain  global representation~\cite{mr,spoc,rmac,crow,interactive,PWA,SBA} and achieve better performance.
Recently, many methods seek to fine-tune the image representations~\cite{netvlad,fine_tune_1,fine_tune_2,CRN,fine_tune_3,fine_tune_4,delf} for image retrieval.
Fine-tuning significantly improves the adaptation ability of the pre-trained network.

\begin{figure}
  \centering
  \includegraphics[width=2.5 in]{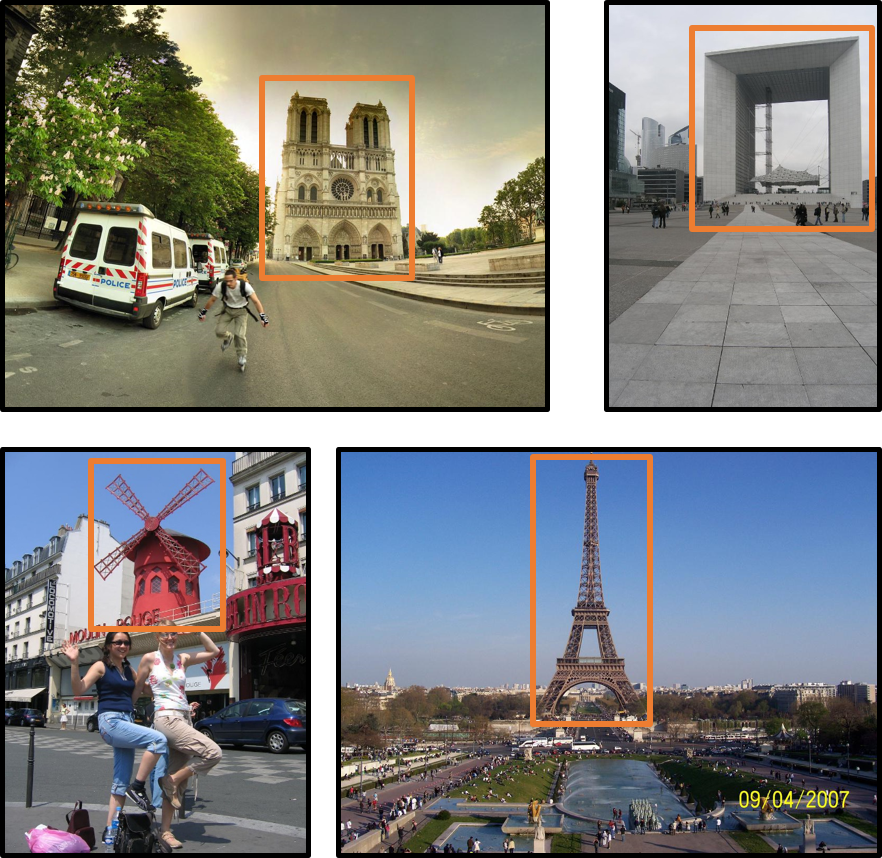}\\
  \caption{Some examples of images that only contain partial interested objects in Paris dataset~\cite{paris}. As marked by orange box, the interested object only occupies a small fraction of the area and is easy to be interfered by the background in image retrieval.
  }
  \label{partial_foreground}
\end{figure}

\begin{figure*}
  \centering
  \includegraphics[width=6.8 in]{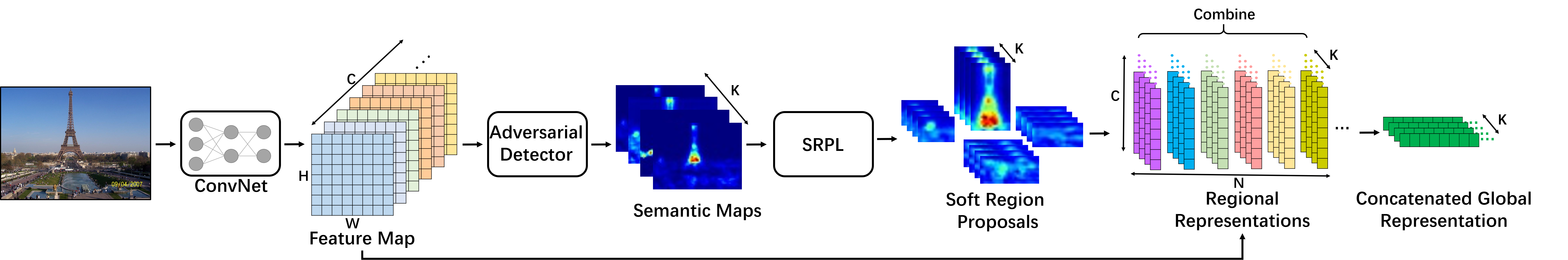}\\
  \caption{The framework of our weakly supervised adversarial soft-detection-based aggregation (ASDA) method.
  Firstly, we extract feature maps of deep convolutional layers and generate some soft region proposals on the feature maps by soft region proposal layer (SRPL).
  Then, we extract regional representations based on generated soft region proposals.
  Finally, we combine regional representations to obtain global representation and concatenate global representations corresponding to various semantic maps to obtain concatenated global representation.
  }
  \label{flow_chart}
\end{figure*}

%检测
%一般特征聚合都是基于全图的
Most of the previous image representations aggregation methods for image retrieval are based on the whole image and ignore the discriminative regional information.
%但是一般目标只是图像的一部分（举例），所以获得关键的regional 的信息有助于更精准的图像检索。
However, a lot of images only contain partial interested objects, as shown in Fig.~\ref{partial_foreground}.
Therefore, it is important to capture the interested region and extract pivotal regional representation in image retrieval.
%rmac和DIR基于检测和局部信息的聚合.
R-MAC~\cite{rmac} utilizes regional information based on the sliding window.
DIR~\cite{fine_tune_2,fine_tune_3} employs region proposal network (RPN) in faster-rcnn~\cite{faster_rcnn} to replace the rigid grid.
%RPN需要框才能训练
Its RPN is trained using the annotated bounding box of the Landmarks dataset, similar to the proposal mechanism in~\cite{faster_rcnn}.
%但是rmac和DIR每个区域重要性一样，没有利用detection score。 （而且DIR 需要基于框的标注信息。）
The regional representation is calculated based on rectangle proposal (named as hard region proposal in this paper), and the significance of each regional representation is uniform in R-MAC~\cite{rmac} and DIR~\cite{fine_tune_2,fine_tune_3}.
Recently, SBA~\cite{PWA,SBA} proposes the soft region proposals generated by the empirical strategy to highlight the certain discriminative pattern.
In order to gain pivotal regional representations, we propose the trainable soft region proposal that can be interpreted as regional attention-based significance in aggregation stage to highlight discriminative semantic contents and patterns.

%检测方法介绍，弱监督策略
In recent years, many methods achieve remarkable performance in object detection, such as Region-based Convolutional Neural Networks (R-CNN)~\cite{rcnn}, Fast R-CNN~\cite{fast_rcnn}, Faster R-CNN~\cite{faster_rcnn}, Mask R-CNN~\cite{mask_rcnn}, You Look Only Once (YOLO)~\cite{yolo},  Single Shot Detector (SSD)~\cite{ssd}, and RetinaNet~\cite{RetinaNet}.
However, these methods are trained based on the annotated bounding box and it is very laborious and expensive to collect bounding box annotations.
By contrast, the image-level annotation is much easier to acquire.
Many weakly supervised object detection and localization methods~\cite{wsddn,wsrpn,loc1,loc2} trained without annotated bounding box are proposed recently.
In order to over tackle the shrinkage problem, some weakly supervised object localization methods~\cite{AE,ACOL} obtain complementary object regions based on adversarial erasing strategy.

Inspired by above previous researches, we propose a weakly supervised adversarial soft-detection-based aggregation (ASDA) method based on trainable soft region proposals for image retrieval in this paper.
We generate semantic maps corresponding to some special discriminative patterns by the proposed adversarial detector.
The soft region proposals based on trainable semantic maps highlight the discriminative pattern of objects and suppress the noise of background.
Different from previous object detection methods~\cite{ssw,eg,rcnn,fast_rcnn,faster_rcnn,yolo,ssd} that use rectangular boxes to represent interested objects, we utilize arbitrarily shaped soft region proposals that indicate the probability of object as the foreground to reflect significance information.
Note, our weakly supervised adversarial soft-detection-based aggregation method is free from annotated bounding box.
We holistically train the detection and aggregation network only based on image-level annotations.

The diagram of the proposed method is illustrated in Fig.~\ref{flow_chart}.
Firstly, we extract feature maps of deep convolutional layers and generate some semantic maps corresponding to various patterns and semantic contents by trainable adversarial detectors.
Then, we generate some initial candidate regions on the feature maps and compute the soft region proposals by soft region proposal layer (SRPL) based on the semantic maps and candidate regions.
Finally, we combine regional representations as global representation and concatenate global representations corresponding to various semantic maps to obtain concatenated global representation.
As far as we know, our adversarial soft-detection-based aggregation method is the first work that introduces weakly supervised adversarial detection into feature aggregation for image retrieval.
%本方法可以使用基于cnn的检测算法，也可以使用传统检测算法。与现存的一般的检测方法都可以结合，可以不使用有监督的目标框信息，仅仅借相似对即可训练。
%The frame of our method can employ both traditional object detection methods (such as Sliding Window (SW), Selective Search Windows (SSW)~\cite{ssw}, and EdgeBoxes (EB)~\cite{eg}) and CNN-based methods(such as R-CNN~\cite{rcnn}, Fast R-CNN~\cite{fast_rcnn}, Faster R-CNN~\cite{faster_rcnn}, Mask R-CNN~\cite{mask_rcnn}).
%with no human ~\cite{fine_tune_4}.
%与聚合更适合的检测算法改进，直接操作深层卷基层的feature map, 计算代价小。
%We directly employ these detection methods on feature map rather than on original image and aggregate the representation based on interested region of feature map.
%Because the region of interest are directly utilized on feature map, the cost of aggregation is lower.
%据我们所知，我们是第一个将弱监督检测结合到图像检索中的特征聚合的。

The main contributions of this paper can be summarized as follows:
\begin{itemize}
\item
We propose a novel trainable adversarial detector to generate semantic maps corresponding to various discriminative patterns and semantic contents.
Benefiting from the adversarial erasing strategy, generated semantic maps preserve more discriminative and detailed information.
\item
Our trainable soft region proposal is arbitrary shape rather than only rectangle and it reflects the significance of objects and various parts of it.
The aggregation based on soft region proposal highlights discriminative semantic contents and suppresses the noise of background.
\end{itemize}

\begin{figure*}
  \centering
  \includegraphics[width=6.8 in]{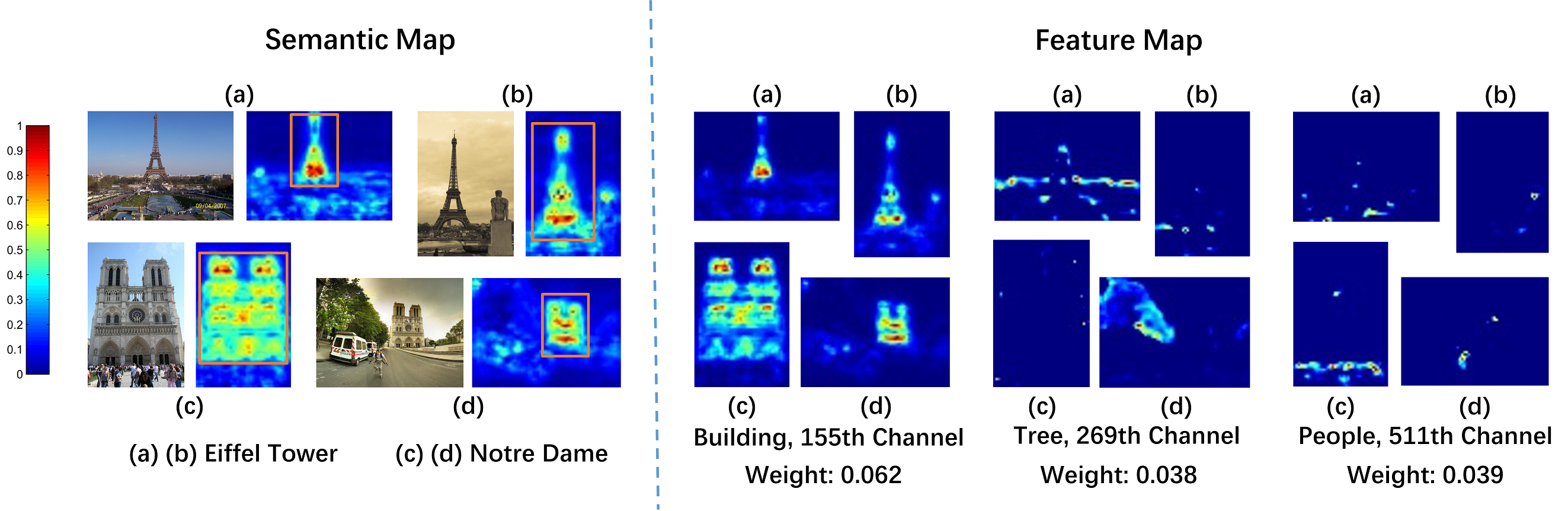}\\
  \caption{Some examples of the semantic map without adversarial learning. Left: semantic map highlights pivotal objects (buildings) and suppresses the noise background (such as people and trees). (a)(b) are Eiffel Tower and (c)(d) are Notre Dame. Right: Some representative channels of the feature map. The important channel is emphasized by larger weight such as 155th channel corresponding to building.
  }
  \label{M}
\end{figure*}

\begin{figure*}
  \centering
  \includegraphics[width=6.8 in]{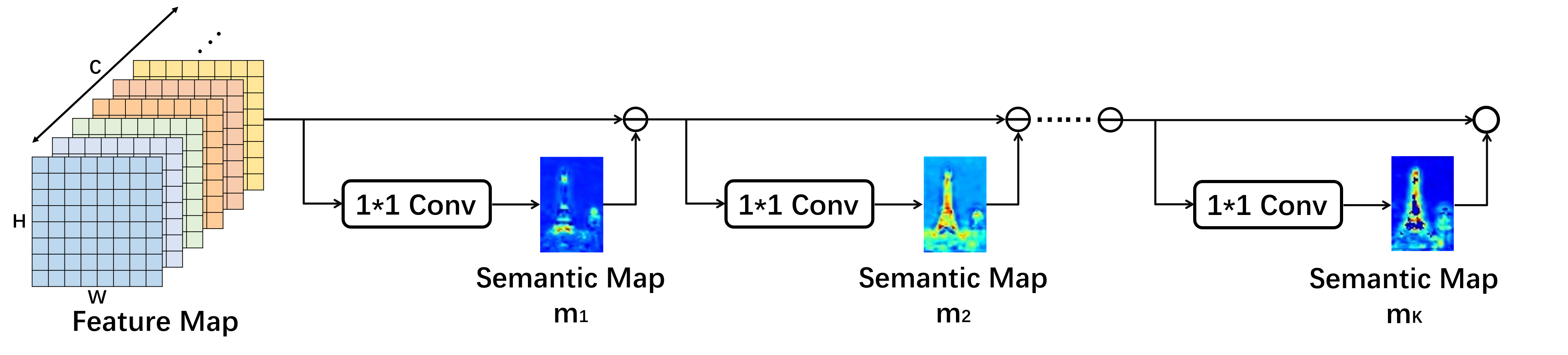}\\
  \caption{Adversarial detector. Our adversarial detector generates some semantic maps ($m_{k}$ ($k=1, 2, \cdots , K$ ) corresponding to special patterns based on adversarial strategy. The $k_{th}$ semantic map $m_{k}$ pays more attention to the semantic content different from previous $k-1$ semantic maps because the activation in previous $k-1$ semantic maps is erased step by step.
  }
  \label{Adversarial detectors}
\end{figure*}

\section{Method}
In this section, we introduce our adversarial soft-detection-based aggregation (ASDA) method in detail.
The framework of the proposed ASDA method is illustrated in Fig~\ref{flow_chart}.
Firstly, generate some semantic maps $m_{k}$ ($k=1, 2, \cdots , K$ , where K is the number of semantic maps) corresponding to special patterns by adversarial detectors.
Then, we obtain initial candidate regions $r_{i}$ ($i=1, 2, \cdots , I$ , where I is the number of candidate regions) based on traditional or CNN-based detection methods, and generate soft region proposals $srp_{i,k}$ based on candidate regions and semantic maps by soft region proposal layer (SRPL).
In the next step, we extract regional representations based on generated soft region proposals which indicate the significance of whole candidate regions and various parts of them.
%We can use Max Pooling (MAC)~\cite{rmac}, Average Pooling (SPoC)~\cite{spoc}, Generalized-Mean Pooling (GeM)~\cite{fine_tune_4}, and CroW~\cite{crow} to extract the regional representations based on activations of deep convolutional layers.
Finally, we combine regional representations $rep_{i,k}$ to obtain global representation $g_{k}$ and concatenate global representations corresponding to various semantic maps to obtain concatenated global representation$g$.
We employ weakly supervised loss that only need annotated tuples $(q,p,n)$ containing query image, positive image that matches the query, and negative image that do not match the query.
It is worth noting that our method is trained without bounding boxes annotations.

\subsection{Adversarial Soft-detection-based Aggregation}
\subsubsection{Adversarial detector}
%加比较sba，需要选择很多semantic detector，造成串联维度太高。我们基于Adversarial 学习，生成互补的semantic detector，仅需要少量的detector。
%l2归一化 平均不同semantic map的贡献，减弱训练不平衡。

%首先介绍 semantic map ，然后介绍基于adversarial学习获得的semantic maps.
\textbf{Semantic map.}
The detector $D()$ generates a semantic map ${\rm{m}} = D(f)$ based on feature map $f$ of deep convolutional layer.
The semantic map represents the significance of each position on the feature map.
%详细介绍 semantic map
%加图片示例，说明 semantic map的作用，详细分析。
In order to explain the semantic map in detail, we illustrate how it works in Fig.~\ref{M}.
The semantic map highlights pivotal objects and suppresses the noise background such as people and trees.
%建筑物被强调， Eiffel Tower和Notre Dame 所在位置 具有较高的值， 对应较高的significance。
The value of positions of buildings such as Eiffel Tower and Notre Dame on the feature map is large, which means the features of these positions are important for discriminability.
By contrast, the trees and people are distractions in this image retrieval task, and they can be suppressed by our semantic map which can be interpreted as a spatial selection.
%在计算Semantic map $M$的过程中，是对各通道的加权挑选，挑选对于训练数据区分更有利的语义内容（可视化不同通道和最后结果）。
The effectiveness and validity of semantic map benefit by $1 \times 1$ convolutional layer, which weights and combines various channels of feature map corresponding to different semantic content as the semantic map.
The important channel is emphasized by a larger weight of $1 \times 1$ convolutional layer and noise is suppressed.
% in the computation of semantic map $M$ based on the feature map of the deep convolutional layer.
For example, the 155th channel of feature map represent building and its weight is 0.062.
The weights of the channels corresponding to the tree (0.038) and people (0.039) are smaller.
As a result, we select some important channels corresponding to discriminative semantic content to generate a semantic map, which reflects the significance of various position.
The impact of our semantic map is similar to the attention-based keypoint selection in DELF~\cite{delf}.
In order to capture more discriminative and detailed semantic contents, we introduce the adversarial erasing strategy into the detector.

\textbf{Adversarial erasing.} The semantic map can be simply generated by a $1 \times 1$ convolutional layer, but only one semantic map can not express enough rich semantic information.
In order to learn the semantic maps that contain different semantic contents, we introduce the adversarial erasing strategy into our adversarial detector.
The overview of the proposed adversarial detector is shown in Fig.~\ref{Adversarial detectors}.
%每一级都生成一个semantic map，k级的semantic map不同于之前的k-1 级语义内容。
Our adversarial detector generates a semantic map in each step.
In order to force the $k_{th}$ detector to capture the different semantic contents from previous $k-1$ semantic maps, we erase the input data stream $f_{k-1}$ of $k_{th}$ detector by the $k-1_{th}$ semantic map.
%加公式
The semantic map $m_{k}$ in step $k$ is calculated by
\begin{equation}\label{5}
{m_k} = {D_k}({f_k})
\end{equation}
Where we define the detector in step $k$ as $D_k$ and the input data stream in step $k$ as $f_k$. The input data stream $f_k$ is calculated based on adversarial erasing strategy.
\begin{equation}\label{6}
{f_k} = {r_{k}}{f_{k - 1}}
\end{equation}
Where $r_{k}=m_{k-1} < \theta$ is the residual region from step $k-1$ to step $k$. $\theta$ define the adversarial threshold to control adversarial intensity.
The special semantic contents in the data stream are erased step by step, and $\ell_{2}$-normalized in Formulation.~\ref{2} balances the contribution of various semantic maps for final representation.
Different from previous adversarial erasing approach~\cite{AE,ACOL}, our adversarial detector is based on holistic representation rather than multiple classifiers.

%比较有对抗生成的semantic maps 和 没有对抗生成的semantic maps。
%底座 塔身 塔外轮廓
As shown in Fig.~\ref{Adversarial detectors}, the adversarial semantic maps indicate different patterns respectively. The semantic map $m_{1}$ highlights the bottom of Eiffel Tower. The semantic map $m_{2}$ focuses on the body of buildings. The semantic map $m_{K}$ highlights the outer contour of Eiffel Tower.
Compared with the semantic map generated without adversarial learning as shown in Fig.~\ref{M}(b), the adversarial detectors capture more discriminative and detailed patterns.

\subsubsection{Soft region proposal layer (SRPL)}
\label{SRPL}

\begin{figure}
  \centering
  \includegraphics[width=3.0 in]{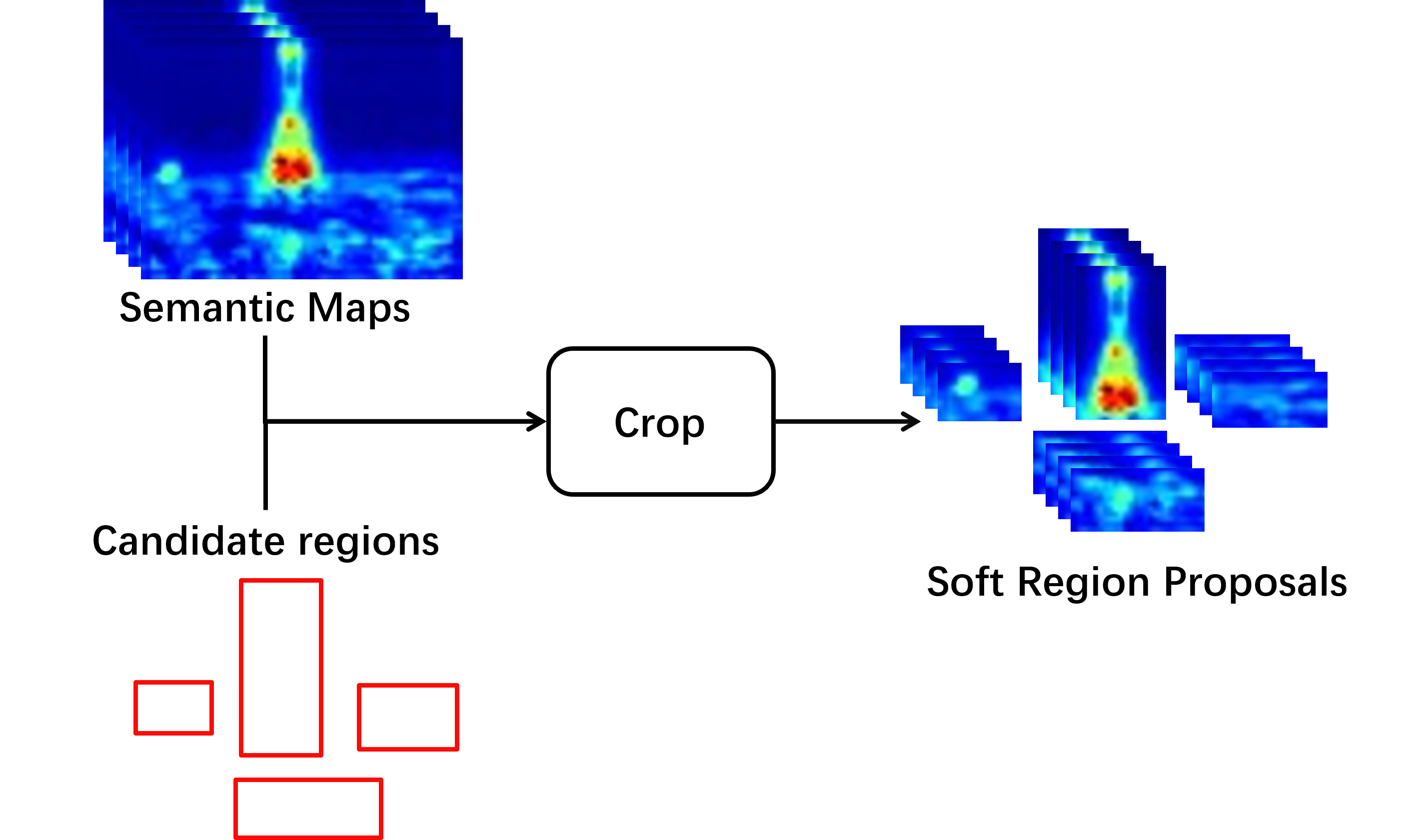}\\
  \caption{Soft region proposal layer (SRPL). We generate soft region proposals based on candidate regions and semantic maps by soft region proposals layer. The initial candidate regions can be obtained by general detection methods, e.g. Multi-scale Sliding Window employed in R-MAC~\cite{rmac}.
  This architecture is naturally implemented with cropping operation on semantic maps generated by adversarial detectors.
  }
  \label{SRPL_image}
\end{figure}

%SRPL 细化介绍SRPL的流程和内容 计算Semantic map 根据 candidate regions 截取
We generate soft region proposals by soft region proposal layer (SRPL).
The overview of the soft region proposals layer is illustrated in Fig.~\ref{SRPL_image}.
%Semantic map $M$ is computed with the input of feature maps of the deep convolutional layer by $1 \times 1$ convolutional kernel, which can be interpreted as channel selection.
The semantic map $m_{k}$ is generated by trainable adversarial detectors, which corresponding to various patterns and semantic contents.
The value of the semantic map reflects the significance of each position on the feature map.
We crop semantic map $m_{k}$ by candidate regions $r_{i}$ to generate soft region proposals $srp_{i,k}$=$Crop(m_{k},r_{i})$.

%检测region
We can generate some initial candidate regions based on general detection methods, such as Multi-scale Sliding Window used in R-MAC~\cite{rmac}.
Different from previous general object detection methods, we pay more attention to the position of candidate regions on the feature maps of deep convolutional layers rather than the precise location on original images.
%candidate regions 是最初始的候选，我们使用 soft region proposal layer 细化获得更有效的区域重要性表示。
The candidate regions work as an initial candidate, then we generate semantic map and soft region proposals to represent the detailed significance of regions.
Therefore, the dependency of the detection precision of initial detection methods is low.
Even simple Sliding Window method can meet the requirements.

%基于 semantic map 使用  candidate box 进行截取 提取 soft region proposals。soft region proposals不局限于矩形，并且可以反映region 中每个空间部分的重要性， 在下一步聚合中起到关键的空间加权作用。
Semantic map $m_{k}$ and initial candidate regions $r_{i}$ are employed to generate soft region proposals $srp_{i,k}$.
We crop semantic map based on the coordinates of candidate boxes, and the regional cropped semantic maps are defined as soft region proposals.
The soft characteristic is mainly embodied in value and shape. The ``soft" value is the fractional probability (the probability represents the significance of interested object) rather than binary (it is the interested object or not). The ``soft" shape is arbitrary shape rather than only rectangle, which benefits by ``soft" value for each position.
Different from previous hard region proposal methods~\cite{rmac,fine_tune_2,fine_tune_3}, our trainable soft region proposal is more flexible and effective for image retrieval.
Compared to SBA~\cite{PWA,SBA} that extract soft region proposals by empirical strategy, our adversarial detectors generate learning-based soft region proposal based on adversarial erasing strategy.
Benefited from adversarial strategy, our soft region proposals are discriminative and low-redundancy.
Therefore, our method just needs a few numbers of soft region proposals compared to SBA~\cite{PWA,SBA} ($K_{ASDA}=4$ in our ASDA and $K_{SBA}=25$ in SBA).

\subsubsection{Aggregation}
\label{Aggregation}
%使我们的Soft-detection-based Aggregation 包含了通道的选择和空间的选择，利用深度卷积神经网络的深层语义信息获得鉴别性的局部representation。
The soft region proposals play an important role in aggregation stage, working as weakly supervised spatial and channel selection.
%局部特征表示
We extract regional representations from feature maps by employing the candidate boxes $r_{i}$ on the feature maps as pooling window and weighting the $H_{i}\times W_{i} \times C$ cropped feature maps $f_{i}$ by corresponding $H_{i}\times W_{i}$ soft region proposals  $srp_{i,k}$ for semantic map $m_{k}$.
%Various pooling methods (e.g. Max Pooling (MAC)~\cite{rmac}, Average Pooling (SPoC)~\cite{spoc}, Generalized-Mean Pooling (GeM)~\cite{fine_tune_4}, and CroW~\cite{crow}) can be employed in our framework.
The C-dimensional regional representations $rep_{i,k}$ for semantic map $m_{k}$ and candidate boxes $r_{i}$ are computed by Max Pooling (MAC)~\cite{rmac} of weighted cropped feature maps as follows:
\begin{equation}\label{1}
rep_{i,k} = MAC\left( {srp_{i,k} * {f_i}} \right)
\end{equation}
Spatially weighted by soft region proposal, the significance of each regional representation for final global representation is adjusted by its intensity implicitly.
%detection score
%Trainable detection scores corresponding to each candidate region indicate the significance of proposals for global representation.
%Detection scores are computed based on corresponding regional representations.
Various other pooling methods (such as Average Pooling (AVG)~\cite{spoc} and Generalized-Mean Pooling (GeM)~\cite{fine_tune_4}) can be employed as a candidate to replace Max Pooling (MAC) in our framework.

%聚合局部representation
Then, we combine regional representations $rep_{i,k}$ to generate global representations $g_{k}$ for semantic map $m_{k}$.
Regional representations are weighted by corresponding soft region proposals  $srp_{i,k}$ to highlight important regions before.
%重要性加权使没有关键信息的regional representation 被忽略， 重要的regional representation 被加强。
The value of regional representation that mainly contains background and noise is small.
By contrast, the value of regional representation that contains interested objects is large.
Therefore, we directly sum the weighted regional representations $rep_{i}$ and $\ell_{2}$-normalized as C-dimensional global representations $g_{k}$ for semantic map $m_{k}$:
\begin{equation}\label{2}
g_{k}{\rm{ = }}{\ell _2}(\sum\limits_{\rm{i}} {rep_{i,k}} )
\end{equation}
Where $\ell_{2}()$ is the $\ell_{2}$-normalized function to normalize representation vector.
The global representation $g_{k}$ for each semantic map is $\ell_{2}$-normalized to balance the contribution of various semantic maps.
The C-dimensional global representations $g_{k}$ for semantic map $m_{k}$ are concatenated as $K \times C$-dimensional concatenated global representation $g$:
\begin{equation}\label{3}
{\rm{g}} = [{g_1},{g_2},...,{g_K}]
\end{equation}

%降维
Finally, we reduce the dimensionality of concatenated global representation $g$ by a fully connected layer and obtain the final D-dimensional compact global representation $rep$, which is $\ell_{2}$-normalized again after dimensionality reduction. $\ell_{2}$-normalized is often employed as standard post-process~\cite{PWA,fine_tune_3,fine_tune_4} in image retrieval.
\begin{equation}\label{4}
rep{\rm{ = }}{\ell _2}(RD(g))
\end{equation}
Where our dimensionality reduction function based on fully connected layer is defined as $RD()$.
This dimensionality reduction strategy is embedded in the network and is trained end-to-end by our weakly supervised loss.
The dimensionality of final representation can be chosen according to the tradeoff between
performance and efficiency in practice.

\subsection{Global Contrastive Loss}
%representation构成中包含region 信息（motivation：希望region使表示更具有鉴别性），损失函数，标记数据
The global representation is aggregated based on regional significance information.
We train the network based on global representation and optimize the significance of regional representation and significance of spatial position and channels end-to-end by contrastive loss~\cite{contrastive_loss}.
The training data consist of image tuple $(q,p,n)$ without candidate boxes annotation, that define query image, positive image that matches the query, and negative image that do not match the query respectively.
We employ the contrastive loss~\cite{contrastive_loss} to force the distance of matching pairs closer than no-matching pairs.
\begin{equation}\label{4}
\begin{split}
{L_{con}}(q,p,n) = & {\left\| {rep(q) - rep(p)} \right\|^2} + \\
 &\max {(0,\tau  - \left\| {rep(q) - rep(n)} \right\|)^2}
\end{split}
\end{equation}
Where $\tau$ is a margin parameter defining when non-matching pairs $(q,n)$ have large enough distance in order not to be taken into account in the loss.
The $\ell_{2}$-normalized  vectors $rep(q)$, $rep(p)$ and $rep(n)$ define D-dimensional global representation for images of tuple $(q,p,n)$ respectively.

\section{Experiments}
\subsection{Datasets}
In this paper, we evaluate the performance of our method on four image retrieval datasets.
The mean average precision (mAP)~\cite{nc,mr,spoc,rmac,crow,PWA,SBA,netvlad,fine_tune_1,fine_tune_2,CRN,fine_tune_3,fine_tune_4} over the query images is employed to evaluate the performance.

Oxford~\cite{oxford} and Paris~\cite{paris} datasets contain photographs collected from Flickr associated with Oxford and Paris landmarks respectively. They consist of 5063 and 6392 images respectively and each dataset contains 55 queries coming from 11 landmarks.

Revisited Oxford (ROxford) and Revisited Paris (RParis)~\cite{ROxford}  alleviate some inaccuracies in the original annotation of Oxford~\cite{oxford} and Paris~\cite{paris} datasets, and introduce new and more challenging 15 queries from 5 out of the original 11 landmarks.  Along with the 55 original queries from Oxford~\cite{oxford} and Paris~\cite{paris}, they consist of a total of 70 queries per dataset.
Three evaluation setups of different difficulty are defined by treating labels (easy, hard, unclear) as positive or negative, or ignoring them.
Easy (E): easy images are treated as positive, while hard and unclear are ignored.
Medium (M): easy and Hard images are treated as positive, while unclear are ignored.
Hard (H): hard images are treated as positive, while easy and unclear are ignored.
The old setup of Oxford~\cite{oxford} and Paris~\cite{paris} appears to be close to the new Easy setup, while Medium and Hard appear to be more challenging.
Because the performance of the Easy setup is similar to old setup and nearly saturated, we do not use it but only evaluate Medium and Hard setups in the subsequent experiments.

\subsection{Implementation Details}
%后处理：ms(放在实验)，白化（也可以放在实验）, 训练集
%pytoch。网络vgg16。 由于显存太大和计算损耗没有使用resent。
%\subsubsection{CNN architectures.}
\textbf{CNN architectures.} We employ VGG16~\cite{VGG} which are highly influential in image retrieval task.
VGG16 network has compromised complexity and performance.
%The performance of ResNet-based methods is better on Paris and RParis datasets.
We do not employ ResNet101~\cite{ResNet} in this paper, due to its insignificant performance improvement with large memory cost of GPU and high computational complexity.
We use PyTorch to implement our proposed method and use Adam~\cite{adam} as the optimizer.
%参数设置
The initial learning rate is $10^{-6}$, an exponential decay is $e^{-0.1i}$ over epoch $i$, momentum is $0.9$, and weight decay is $5\times10^{-4}$.

\textbf{Improve efficiency.}
%为了加速 我们直接求一次 与feature map 相乘.只截取一次。我们先加权后截取
In order to improve computational efficiency,  we spatially weight feature map by semantic map and crop the weighted feature map based on candidate regions $r_{i}$ directly in practice.
The cropped weighted feature maps are aggregated by Max Pooling (MAC) to generate C-dimensional regional representations.
In this way, we only weight once for all candidate regions rather than once for each candidate region, and only crop once for each candidate region on spatially weighted feature map rather than twice for each candidate region on feature map and semantic map respectively.
%解释 practical method的理解
Our practical method can be interpreted as the attention-based strategy, which consists of channel selection and spatial selection.
Different from previous region-based methods~\cite{rmac,fine_tune_2,fine_tune_3}, our practical method predicts the significance of each region and embed it in corresponding regional representation without $\ell_{2}$-normalized operation.

%ms
%\subsubsection{Multi-scale (MS).}
\textbf{Multi-scale (MS).}
During test time we adopt a multi-scale procedure similar to the recent works~\cite{fine_tune_3,fine_tune_4}, which improves the performance in image retrieval.
We resize the input image to different sizes and feed multiple input images of various scales $(1, 1 / \sqrt{2} , and~1 / 2 )$ to the network.
Then, the global representations from multiple scales are combined into a single representation.

%白化
%\subsubsection{Learned whitening (LW).}
\textbf{Learned whitening (LW).}
Previous unsupervised aggregation methods~\cite{spoc, PWA} generally employ PCA of an independent set for whitening and dimensionality reduction as post-processing.
Based on labeled data, we employ linear discriminant projections~\cite{lw} used in recent works~\cite{fine_tune_1,fine_tune_4} to project the representations.

%训练集
%\subsubsection{Training data.}
\textbf{Training data.}
The training dataset is collected and used in~\cite{fine_tune_4}. It is labeled based on Structure-from-Motion (SfM) with no human annotation.
There are around 133k images for training and around 30k images for validation.
Each training and validation tuple contains 1 query, 1 positive and 5 negative images.
We set the margin for contrastive loss as $\tau=0.75$ and batch size of training tuples as 5.

%detection method
%\subsubsection{Detection.}
\textbf{Adversarial detector.}
We use a simple object detection method Sliding Window~\cite{rmac} to generate candidate regions. We slide windows of $L$  scales on feature maps and the overlap of neighboring regions is small than 0.4.
The possible region number for the long dimension is (2, 3, 4, 5, 6) at scale (1, 2, 3, 4, 5) respectively.
The aspect ratio is simply set as 1:1.
The number of semantic maps is set as $K=4$ and the adversarial threshold is set as $\theta=0.7$.
Query objects are not cropped based on the boxes provided by the datasets in this paper because our adversarial detector can capture the pivotal object.

\subsection{Impact of the Parameters}
We evaluate the main parameters and processes that affect the performance of the proposed adversarial soft-detection-based aggregation (ASDA) method in this section.
\subsubsection{Scale number}
%region num  of  slide windows
We employ the simple initial detection method Sliding Window in our method to generate initial candidate boxes.
The scale number ($L$) of slide windows are evaluated as shown in Table.~\ref{scale}.
We only use full-sized representation when $L=0$.
%L=4时效果较好，可以检索出较小的目标。
Employing slide windows of 4 scales, we achieve the best performance.
The small-sized initial candidate boxes are beneficial for small object retrieval.
Therefore, the improvement of image retrieval performance is more significant in Paris dataset which contains many small query objects as shown in Fig.~\ref{partial_foreground}.

\begin{table}[htbp]
  \centering
  \caption{Performance (mAP) of varying scale number ($L$) of slide windows. Only full-sized representation is utilized when $L=0$.
  }
   %\scalebox{1.5}{
    \vspace{0.5em}
    \begin{tabular}{ccc}
    \toprule[1.25pt]
    %\hline
          & \multicolumn{2}{c}{\textbf{Datasets}} \bigstrut\\
\cline{2-3}    \multicolumn{1}{c}{\textbf{$L$}} & \multicolumn{1}{c}{Oxford} & \multicolumn{1}{c}{Paris} \bigstrut\\
    \hline
    \multicolumn{1}{c}{0} & \multicolumn{1}{c}{81.5} & \multicolumn{1}{c}{81.1} \bigstrut[t]\\
    \multicolumn{1}{c}{1} & \multicolumn{1}{c}{82.4} & \multicolumn{1}{c}{81.9} \\
    \multicolumn{1}{c}{2} & \multicolumn{1}{c}{83.1} & \multicolumn{1}{c}{83.4} \\
    \multicolumn{1}{c}{3} & \multicolumn{1}{c}{83.1} & \multicolumn{1}{c}{84.8} \\
    \multicolumn{1}{c}{4} & \multicolumn{1}{c}{\textbf{83.5}} & \multicolumn{1}{c}{\textbf{85.4}} \\
    \multicolumn{1}{c}{5} & \multicolumn{1}{c}{83.1} & \multicolumn{1}{c}{85.1} \bigstrut[b]\\
    %\hline
    \bottomrule[1.25pt]
    \end{tabular}%
   %}
  \label{scale}%
\end{table}%

\subsubsection{Dimensionality}
%维度
In order to get shorter representations, we compress the C-dimensional ($C$=512) global representation $g$ by fully connected layer.
The dimensionality ($D$) of final compact $\ell_{2}$-normalized representation changes from 512 to 32, and the performance of representations with varying dimensionality is reported in Table~\ref{Dim}.
The results show that the performance declines gradually with a decrease of dimensionality.
The dimensionality of final representation can be chosen according to the tradeoff between performance and efficiency in practice.

\begin{table}[htbp]
  \centering
  \caption{Performance (mAP) of varying dimensionality ($D$) of final compact $\ell_{2}$-normalized representation.
  }
   %\scalebox{1.5}{
    \vspace{0.5em}
    \begin{tabular}{ccc}
    \toprule[1.25pt]
    %\hline
          & \multicolumn{2}{c}{\textbf{Datasets}} \bigstrut\\
\cline{2-3}    \multicolumn{1}{c}{\textbf{Dim}} & \multicolumn{1}{c}{Oxford} & \multicolumn{1}{c}{Paris} \bigstrut\\
    \hline
    \multicolumn{1}{c}{32} & \multicolumn{1}{c}{65.5} & \multicolumn{1}{c}{69.7} \bigstrut[t]\\
    \multicolumn{1}{c}{64} & \multicolumn{1}{c}{74.2} & \multicolumn{1}{c}{78.1} \\
    \multicolumn{1}{c}{128} & \multicolumn{1}{c}{79.1} & \multicolumn{1}{c}{83.1} \\
    \multicolumn{1}{c}{256} & \multicolumn{1}{c}{82.4} & \multicolumn{1}{c}{84.3} \\
    \multicolumn{1}{c}{512} & \multicolumn{1}{c}{\textbf{83.5}} & \multicolumn{1}{c}{\textbf{85.4}} \bigstrut[b]\\
    %\hline
    \bottomrule[1.25pt]
    \end{tabular}%
   %}
  \label{Dim}%
\end{table}%

\subsubsection{Adversarial soft region proposal}
In order to demonstrate the effectiveness of our adversarial detector and soft region proposal, we compare hard region proposal(HDA), soft region proposal(SDA) and adversarial soft region proposal(ASDA) in Table.~\ref{adversarial}.
The result shows that the approach based on the adversarial detector and soft region proposal achieves better performance.
Our adversarial detector captures various semantic contents and soft region proposal preserves more discriminative information for representation.

\begin{table}[htbp]
  \centering
  \caption{Performance (mAP) with different region proposal strategies.
  }
   %\scalebox{1.5}{
    \vspace{0.5em}
    \begin{tabular}{ccc}
    \toprule[1.25pt]
    %\hline
          & \multicolumn{2}{c}{\textbf{Datasets}} \bigstrut\\
\cline{2-3}    \multicolumn{1}{c}{\textbf{Methods}} & \multicolumn{1}{c}{Oxford} & \multicolumn{1}{c}{Paris} \bigstrut\\
    \hline
    \multicolumn{1}{c}{HDA} & \multicolumn{1}{c}{82.1} & \multicolumn{1}{c}{85.4} \bigstrut[t]\\
    \multicolumn{1}{c}{SDA} & \multicolumn{1}{c}{83.5} & \multicolumn{1}{c}{85.4} \\
    \multicolumn{1}{c}{ASDA} & \multicolumn{1}{c}{\textbf{84.2}} & \multicolumn{1}{c}{\textbf{86.2}} \bigstrut[b]\\
    %\hline
    \bottomrule[1.25pt]
    \end{tabular}%
   %}
  \label{adversarial}%
\end{table}%

%不同的聚合方式
\subsubsection{Pooling strategy}
In our framework, we can employ various pooling strategies to aggregate weighted cropped feature maps and obtain regional representations.
We show the results of different pooling strategies such as Max Pooling (MAC)~\cite{rmac}, Average Pooling (AVG)~\cite{spoc} and Generalized-Mean Pooling (GeM)~\cite{fine_tune_4} in Table~\ref{pooling}.
The experimental results show that the performance of Max Pooling (MAC)~\cite{rmac} in our adversarial soft-detection-based aggregation method is better than other pooling strategies.

\begin{table}[htbp]
  \centering
  \caption{Performance (mAP) with different pooling strategies.
  }
   %\scalebox{1.5}{
    \vspace{0.5em}
    \begin{tabular}{ccc}
    \toprule[1.25pt]
    %\hline
          & \multicolumn{2}{c}{\textbf{Datasets}} \bigstrut\\
\cline{2-3}    \multicolumn{1}{c}{\textbf{Pooling strategy}} & \multicolumn{1}{c}{Oxford} & \multicolumn{1}{c}{Paris} \bigstrut\\
    \hline
    \multicolumn{1}{c}{AVG} & \multicolumn{1}{c}{81.3} & \multicolumn{1}{c}{84.0} \bigstrut[t]\\
    \multicolumn{1}{c}{GEM} & \multicolumn{1}{c}{83.3} & \multicolumn{1}{c}{85.4} \\
    \multicolumn{1}{c}{MAC} & \multicolumn{1}{c}{\textbf{84.2}} & \multicolumn{1}{c}{\textbf{86.2}} \bigstrut[b]\\
    %\hline
    \bottomrule[1.25pt]
    \end{tabular}%
   %}
  \label{pooling}%
\end{table}%

\subsubsection{Post-process}
%SS MS Lw
We report the performance of original representation without post-process (single-scale, SS) and representation with multi-scale (MS) and learned whitening (LW) post-process in Table.~\ref{process}.
Similarly to the conclusion of recent researches~\cite{fine_tune_3,fine_tune_4}, the results show that post-process such as multi-scale (MS) and learned whitening (LW) can boost the performance of image retrieval.
Therefore, multi-scale (MS) and learned whitening (LW) are generally employed as post-process for image retrieval recently.

\begin{table}[htbp]
  \centering
  \caption{Performance (mAP) with post-process.
  }
   %\scalebox{1.5}{
    \vspace{0.5em}
    \begin{tabular}{ccc}
    \toprule[1.25pt]
    %\hline
          & \multicolumn{2}{c}{\textbf{Datasets}} \bigstrut\\
\cline{2-3}    \multicolumn{1}{c}{\textbf{Post-process}} & \multicolumn{1}{c}{Oxford} & \multicolumn{1}{c}{Paris} \bigstrut\\
    \hline
    \multicolumn{1}{c}{ASDA(SS)} & \multicolumn{1}{c}{84.2} & \multicolumn{1}{c}{86.2} \bigstrut[t]\\
    %\multicolumn{1}{c}{SS+LW} & \multicolumn{1}{c}{85.3} & \multicolumn{1}{c}{88.1} \\
    \multicolumn{1}{c}{ASDA(MS+LW)} & \multicolumn{1}{c}{\textbf{87.7}} & \multicolumn{1}{c}{\textbf{89.0}} \bigstrut[b]\\
    %\hline
    \bottomrule[1.25pt]
    \end{tabular}%
   %}
  \label{process}%
\end{table}%

\begin{table*}[htbp]
  \centering
  \caption{Performance (mAP) comparison with  the state-of-the-art image retrieval methods. Our weakly supervised adversarial soft-detection-based aggregation (ASDA) method outperforms the state-of-the-art aggregation methods on most datasets.
  Especially on the difficult image retrieval datasets ROxford and RParis~\cite{ROxford} that released recently, we achieve significantly boost (3$\%$ $\sim$ 5$\%$).
  The experimental results demonstrate that our ASDA method is effective for image retrieval.
  }
    \vspace{0.5em}
    \begin{tabular}{rrrrrrrr}
     \toprule[1.25pt]
     &     &       \multicolumn{6}{c}{\textbf{Datasets}} \\
\cline{3-8}    \multicolumn{1}{l}{\textbf{Methods}} & \multicolumn{1}{c}{\textbf{Fine-tuned}} & \multicolumn{1}{c}{Oxford} & \multicolumn{1}{c}{Paris} & \multicolumn{2}{c}{ROxford} & \multicolumn{2}{c}{RParis} \\
\cline{5-8}
&  &  &  &  \multicolumn{1}{c}{\textbf{M}}& \multicolumn{1}{c}{\textbf{H}} & \multicolumn{1}{c}{\textbf{M}}& \multicolumn{1}{c}{\textbf{H}}\\
    \toprule[1pt]
    \multicolumn{1}{l}{MAC~\cite{mr}}& \multicolumn{1}{c}{No} & \multicolumn{1}{c}{56.4} & \multicolumn{1}{c}{72.3}  & \multicolumn{1}{c}{37.8} & \multicolumn{1}{c}{14.6}  & \multicolumn{1}{c}{59.2} & \multicolumn{1}{c}{35.9}\\
     \multicolumn{1}{l}{SPoC~\cite{spoc}}& \multicolumn{1}{c}{No}  & \multicolumn{1}{c}{68.1} & \multicolumn{1}{c}{78.2}  & \multicolumn{1}{c}{38.0} & \multicolumn{1}{c}{11.4}  & \multicolumn{1}{c}{59.8} & \multicolumn{1}{c}{32.4}\\
   \multicolumn{1}{l}{CroW~\cite{crow}}& \multicolumn{1}{c}{No}  & \multicolumn{1}{c}{70.8} & \multicolumn{1}{c}{79.7}  & \multicolumn{1}{c}{41.4} & \multicolumn{1}{c}{13.9}  & \multicolumn{1}{c}{62.9} & \multicolumn{1}{c}{36.9}\\
   \multicolumn{1}{l}{R-MAC~\cite{rmac}}& \multicolumn{1}{c}{No}  & \multicolumn{1}{c}{66.9} & \multicolumn{1}{c}{83.0}  & \multicolumn{1}{c}{42.5} & \multicolumn{1}{c}{12.0}  & \multicolumn{1}{c}{66.2} & \multicolumn{1}{c}{40.9}\\
    \multicolumn{1}{l}{NetVLAD~\cite{netvlad}}& \multicolumn{1}{c}{Yes}  & \multicolumn{1}{c}{67.6} & \multicolumn{1}{c}{74.9}  & \multicolumn{1}{c}{37.1} & \multicolumn{1}{c}{13.8}  & \multicolumn{1}{c}{59.8} & \multicolumn{1}{c}{35.0}\\
    \multicolumn{1}{l}{R-MAC~\cite{fine_tune_2}}& \multicolumn{1}{c}{Yes}  & \multicolumn{1}{c}{83.1} & \multicolumn{1}{c}{87.1} & \multicolumn{1}{c}{---} & \multicolumn{1}{c}{---}  & \multicolumn{1}{c}{---} & \multicolumn{1}{c}{---}\\
    \multicolumn{1}{l}{GeM~\cite{fine_tune_4}}& \multicolumn{1}{c}{Yes}  & \multicolumn{1}{c}{\textbf{87.9}} & \multicolumn{1}{c}{87.7}  & \multicolumn{1}{c}{61.9} & \multicolumn{1}{c}{33.7}  & \multicolumn{1}{c}{69.3} & \multicolumn{1}{c}{44.3}\\
    \multicolumn{1}{l}{ASDA}& \multicolumn{1}{c}{Yes}  & \multicolumn{1}{c}{87.7} & \multicolumn{1}{c}{\textbf{89.0}}  & \multicolumn{1}{c}{\textbf{66.4}} & \multicolumn{1}{c}{\textbf{38.5}}  & \multicolumn{1}{c}{\textbf{71.6}} & \multicolumn{1}{c}{\textbf{47.9}}\\
    \toprule[1.25pt]
    \end{tabular}%
  \label{fine-tuned}%
\end{table*}%

\subsection{Comparison With the State-of-the-art}
In the Table.~\ref{fine-tuned}, we extensively compare our adversarial soft-detection-based aggregation (ASDA) method with the state-of-the-art image retrieval methods based on compact image representations.
%In the first part of Table.~\ref{fine-tuned}, the methods are based on VGG~\cite{VGG} network.
The proposed ASDA method based on VGG~\cite{VGG} network outperforms the state-of-the-art on most datasets and achieves comparable performance (only 0.3$\%$, and its result based on Pytorch is 87.2$\%$ on Oxford dataset which is lower than our result (87.7$\%$) also based on Pytorch) on Oxford dataset.
%尤其是较难在最近提出的较难的ROxford和 RParis 有5%的提升。
Especially on the difficult image retrieval datasets ROxford and RParis~\cite{ROxford} that released recently, we achieve significantly boost (3$\%$ $\sim$ 5$\%$).
%因为我们捕获了鉴别性的局部特征，抑制了背景噪声，对于遮挡，小目标等困难样本有较好的效果（可以找一些结果显示一下）。
Because our adversarial soft-detection-based aggregation method exploits the discriminative detailed information by adversarial detector and soft region proposals, which highlight the regional interested objects and suppress the disturb of complex background as illustrated in Fig.~\ref{M},~\ref{Adversarial detectors}.
Adversarial detector captures several semantic maps corresponding to various patterns of object, which preserves more discriminative detail in final compact representations.
Especially for the difficult small query object which is easily interfered by noise of background, our soft region proposals can detect small objects under complex background to promote the discrimination of representation.
The suppressing effect of noise from our trainable soft region proposals is beneficial for representation to overcome the occlusion and disturb of background (such as people and plants).
The experimental results demonstrate that our ASDA method is effective for image retrieval.

%%分析resnet正确率
%%In the first part of  Table III。
%We compare our ASDA method with the state-of-the-art image retrieval methods based on Resnet101 network in the second part of Table.~\ref{fine-tuned}.
%%We achieve better performance on most datasets.
%%paris resnet较差
%We have the state-of-the-art score on ROxford dataset, while our method is outperformed by the recent work~\cite{fine_tune_3} on Paris and RParis datasets with ResNet network.
%Furthermore, it is worth noting that we do not perform any manual labeling or cleaning of our training data (same as~\cite{fine_tune_4}) and do not require annotations of bounding boxes, while in ~\cite{fine_tune_3} landmark labels and annotated bounding boxes are involved.
%%resnet 在paris效果较好 ， 但是 resnet 的计算损耗和内存损耗较大。
%The results show that Resnet-based methods achieve remarkable performance on Paris and RParis datasets.
%However, they achieve insignificant performance improvement on Oxford and ROxford datasets with larger memory cost of GPU and higher computational complexity than VGG-based methods.

\begin{figure}
  \centering
  \includegraphics[width=3 in]{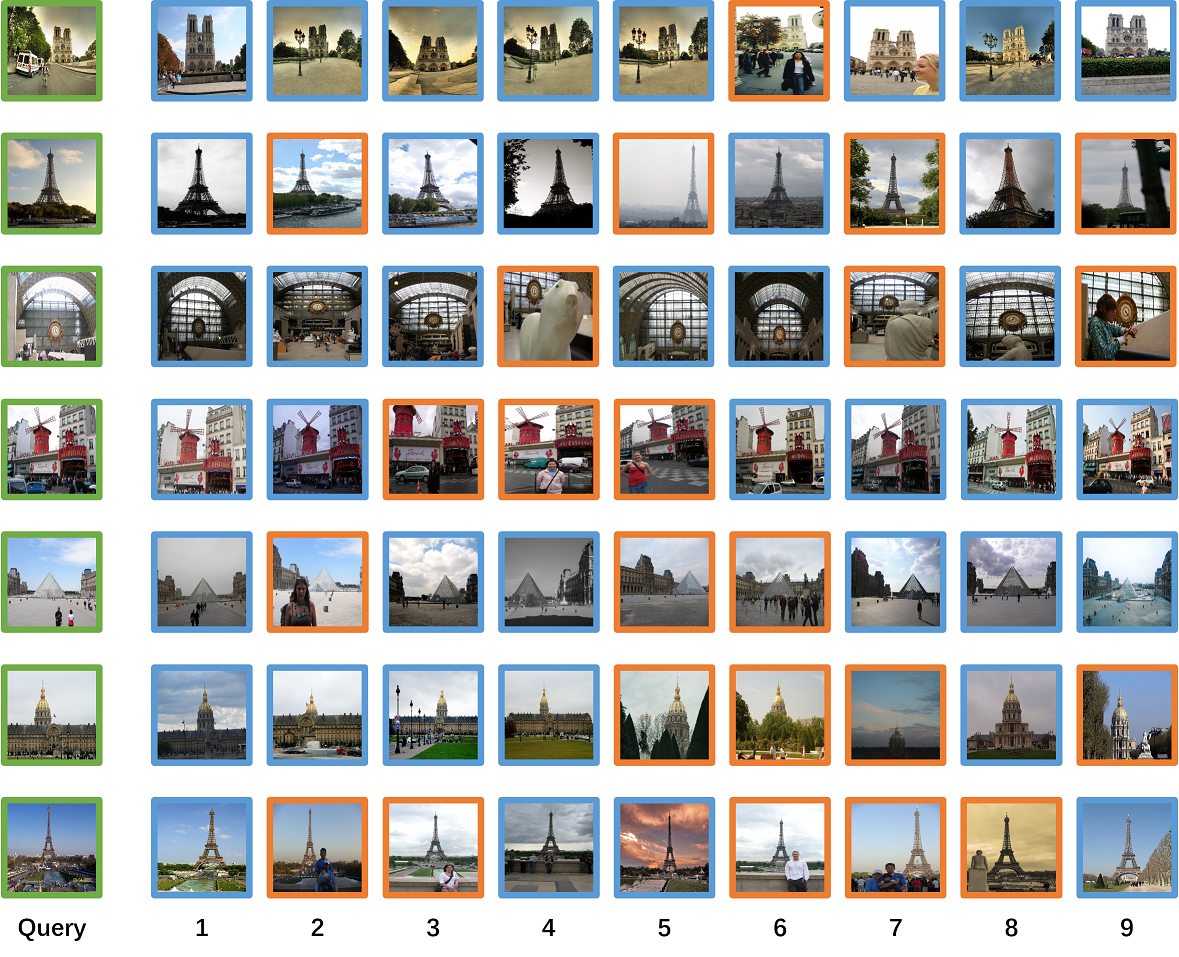}\\
  \caption{Some typical results on Paris dataset. The query images are marked by green boxes.
  The top 9 retrieval results are marked by blue or orange boxes.
  Some hard retrieval results are marked by orange boxes, which contains the occlusion, disturb of people and plants, small objects.
  Our adversarial soft-detection-based aggregation (ASDA) method solves above problems to some extent by trainable channel and spatial selection.
  }
  \label{image_result}
\end{figure}

In order to analyze the results of our method clearly, we show some typical retrieval results in Fig.~\ref{image_result}.
The query images are marked by green boxes.
The top 9 retrieval results are marked by blue or orange boxes.
The orange boxes mark hard candidates, which contains the occlusion, disturb of people and plants, small objects.
The pivotal interested objects that are significant for distinction are highlighted by soft region proposals.
Meanwhile, the noise of background which is harmful for discriminability is suppressed by soft region proposals.
Our adversarial soft-detection-based aggregation method tackles above issue by capturing discriminative regional representations based on soft region proposals and combines them based on predicted significance.
As a result, the final global representation is discriminative and robust.

\section{Conclusion}
In this paper, we propose a novel weakly supervised adversarial soft-detection-based aggregation (ASDA) method for image retrieval.
The key characteristic of our method is that our adversarial detector preserves more discriminative and low-redundancy information for semantic maps and soft region proposals highlight the regional interested objects and suppress the noise of background.
%We predict the significance of different regional representations based on soft region proposals and combine them as compact global representations.
The results show that our method can capture the discriminative information of pivotal small objects without bounding box annotations and overcome the occlusion and disturb of background.

Experiments on some standard retrieval datasets show that our weakly supervised approach outperforms the previous state-of-the-art aggregation methods.
The experimental results demonstrate that the proposed method is effective for image retrieval.
It is worth noting that our weakly supervised  method is trained without bounding box annotations and our soft-detection and aggregation stages are holistically trained end-to-end.
For the future researches of image retrieval, person re-identification and place recognition task, it is worth paying attention to our adversarial soft-detection-based aggregation method that introduces adversarial detector and trainable soft region proposal into image representation learning.

{\small
\bibliographystyle{ieee}
\bibliography{egbib}
}

\end{document}